\documentclass{article}
\usepackage[utf8]{inputenc}
\usepackage[english]{babel}
\usepackage{amssymb}
\newcommand{\cmark}{Yes}%
\newcommand{\xmark}{No}
\usepackage[a4paper,top=3cm,bottom=2cm,left=3cm,right=3cm,marginparwidth=1.75cm]{geometry}

\usepackage{authblk}
\usepackage{blindtext}
\usepackage{amsmath}
\usepackage{graphicx}
\usepackage{multirow}
\usepackage[colorlinks=true, allcolors=blue]{hyperref}
\usepackage{calc}  
\usepackage{array} 
\newlength\mylen
\title{Developing an Effective and Automated Patient Engagement Estimator for Telehealth: A Machine Learning Approach}

\author[1]{Pooja Guhan$^\ddagger$}
\author[1]{Naman Awasthi}
\author[1]{Kathryn McDonald}
\author[2]{Kristin Bussell}
\author[2]{\\Gloria Reeves}
\author[1]{Dinesh Manocha}
\author[3]{Aniket Bera}
\affil[1]{University of Maryland College Park}
\affil[ ]{\textit {\{pguhan, nawasthi, dmanocha\}@umd.edu}}
\affil[2]{University of Maryland, Baltimore County}
\affil[ ]{\textit {\{KMcdonald, kristin.bussell, greeves\}@som.umaryland.edu}}
\affil[3]{Purdue University, West Lafayette, USA}
\affil[ ]{\textit {ab@cs.purdue.edu}}
\date{}
\usepackage{array}
\usepackage{xcolor}

\newcommand{\modelname}{MET}
\newcommand{\dataname}{MEDICA}
\usepackage{multirow, makecell}
\usepackage{subcaption}
\usepackage{booktabs}
\usepackage{amsmath}
\usepackage{amssymb}
\usepackage{amsfonts}
\usepackage{enumitem}
\usepackage{bbm}
\usepackage{dsfont}
\begin{document}

\maketitle

\section{Abstract}

\paragraph{Background:} Patient engagement is a critical but challenging public health priority in behavioral healthcare. During telehealth sessions, healthcare providers need to rely more on verbal strategies than typical non-verbal cues to engage patients. Hence, the typical patient engagement behaviors are now different, and provider training on telehealth patient engagement is unavailable or quite limited. Therefore, we explore the application of machine learning for estimating patient engagement to assist psychotherapists in better diagnosis of mental disorders during telemental health sessions.

\paragraph{Objective:} The objective of this study was to examine the ability of machine learning models to estimate patient engagement levels during a telemental health session and understand whether the machine learning approach could support mental disorder diagnosis by psychotherapists.

\paragraph{Methods:} We propose a multimodal learning-based framework,~\modelname. We uniquely leverage latent vectors corresponding to Affective and Cognitive features frequently used in psychology literature to understand a person’s level of engagement. Given the labeled data constraints that exist in healthcare, we explore a semi-supervised solution using GANs. To further the development of similar technologies that can be useful for telehealth, we also plan to release a dataset MEDICA containing 1299 video clips, each 3 seconds long and show experiments on the same. The efficacy of our method is also demonstrated through real-world experiments.

\paragraph{Results:} Our framework reports a 40\% improvement in RMSE (Root Mean Squared Error) over state-of-the-art methods for engagement estimation. In our real-world tests, we also observed positive correlations between the working alliance inventory scores reported by psychotherapists. This indicates the potential of the proposed model to present patient engagement estimations that aligns well with the engagement measures used by psychotherapists.

\paragraph{Conclusion:} The performance of the framework described here has been compared against other existing engagement detection machine learning models. We also validated the model using a limited sample of real-world data. Patient engagement in literature has been identified to be important to improve therapeutic alliance. But little research has been undertaken to measure it in a telehealth setting wherein the conventional cues are not available to the therapist to take a confident decision. The framework developed is an attempt to model person-oriented engagement modeling theories within machine learning frameworks to estimate the level of engagement of the patient accurately and reliably in telehealth. The results are encouraging and emphasize the value of combining psychology and machine learning to understand patient engagement. Further testing in actual telehealth settings is necessary to fully assess its usefulness in helping therapists gauge patient engagement during virtual sessions. However, the proposed approach and the creation of the new dataset, MEDICA, opens avenues for future research and development of impactful tools for telehealth.



\paragraph{KEYWORDS}
Machine learning, mental health, telehealth, engagement detection, patient engagement

\section{Introduction}
\subsection{Overview}
The World Health Organization defines mental health as “a state of well-being” that allows a person to lead a fulfilling and productive life and contribute to society~\cite{who_mh}. With increasing stress and pressure leading to poor mental health, improved telemental healthcare is becoming a need of the hour as they serve as an effective way to get access to mental health services and treatment in all countries and cultures across the globe. \cite{world2001world}~estimated that one-fourth of the adult population is affected by some kind of mental disorder. However, there are only approximately 9 psychiatrists per 100,000 people in developed countries and only around 0.1 for every 1,000,000 in lower-income countries~\cite{murray2012disability, oladeji2016brain}. Therefore, it is not surprising that there has been an upward trend in the demand for telemental health (the process of providing psychotherapy remotely, typically utilizing HIPAA-compliant video conferencing)~\cite{telemental}~to address the chronic shortage of psychotherapists. These services eliminate some practical barriers to care (e.g., transportation), are affordable, and give access to an actual therapist. Despite these undeniable benefits, this emerging treatment modality raises new challenges in patient engagement compared to in-person care. By engagement, we refer to the connection between a therapist and patient that includes a sense of basic trust and willingness/interest to collaborate which is essential for the therapeutic process. Patient engagement is a critical but challenging public health priority in behavioral health care. There are no objective measurements of patient engagement in behavioral health care. Measurement of engagement is most commonly assessed by patient reports, which may be prone to response bias, and the variable use of different questionnaires makes it challenging to compare patient engagement across different health systems. Behavioral health services often require more frequent appointments than other specialties to promote behavior change, so maintaining a positive relationship with a provider is essential for evidence-based care. However, patient engagement is not routinely or systematically measured in healthcare settings. Health systems often use “show rate” and “patient satisfaction” as a proxy for engagement, but these terms do not necessarily reflect provider-patient alliance in treatment.

In telehealth appointments, therapists have limited visual data (e.g. the therapist can only view the patient’s face rather than their full body). They must rely more on verbal strategies to engage patients than in-person care since they cannot use typical non-verbal cues to convey interest and be responsive to the patient (e.g., handshake at the beginning of a session, adjusting the distance between the patient and provider by moving a chair closer or further away, observing a patient’s response to questions while maintaining eye contact). It is also more difficult for therapists to convey attentiveness since eye contact requires the therapist to look at a camera rather than observing or looking at a person. Additionally, provider training on telehealth patient engagement is quite limited. Providers are currently implementing telehealth services without having clear guidance on how to improve or measure patient telehealth engagement. For example, the abrupt transition to virtual care to prevent COVID-19 transmission did not allow providers to receive training on the use of technology-based care beyond basic orientation to web-based platforms.

Thus, systems that can provide feedback on engagement, using multi-modalities of data, have the potential to improve therapeutic outcomes. Engagement is critical for both retention in care as well as the accuracy of diagnoses. These two factors are potential targets to enhance the quality of technology-delivered care. Therefore, developing a system that can provide feedback on engagement using multimodal data has the potential to improve therapeutic outcomes while performing telemental health.

\subsection{Research Background}
Patient engagement has been established as one of the critical indicators of a successful therapy session. The existing literature in this space largely explores ways of improving it. However, methods to measure or quantify the levels of patient engagement, especially in telehealth settings remain largely unexplored. Some of the prior works in the realm of engagement detection consider using just facial expressions~\cite{whitehill2014faces, murshed2019engagement}, speech~\cite{yu2004detecting}, body posture~\cite{sanghvi2011automatic}, gaze direction~\cite{nakano2010estimating} and head pose  ~\cite{sharma2019student} have been used as single modalities for detecting engagement. Combining different modalities has been observed to improve engagement detection accuracy ~\cite{psaltis2017multimodal, grafsgaard2013embodied, aslan2014learner}. ~\cite{frank2016engagement} proposed a multimodal framework to detect the level of engagement of participants during project meetings in a work environment. The authors expanded the work of Stanford's PBL Labs,  eRing~\cite{ma2015ering}, by including information streams such as facial expressions, voice, and other biometric data.~\cite{monkaresi2016automated} proposed an approach to detect engagement levels in students during a writing task by not only making use of facial features but also features obtained from remote video-based detection of heart rate. The dataset used was generated by the authors, and they used self-reports instead of external annotation for classification purposes.~\cite{chang2018ensemble} make use of facial expressions as well as body posture for detecting engagement in learners. ~\cite{fedotov2018multimodal} proposes the use of audio, facial, and body pose features to detect engagement and disengagement for an imbalanced in-the-wild dataset. 

Despite the existence of a variety of such algorithms to perform engagement detection, the results obtained from these approaches (especially single modality based) could be misleading in a telehealth setting due to factors like camera position, resistant or guarded clients etc. The multimodal architectures have atleast one modality that requires data which cannot be reliably represented or collected. For instance, in video conference calls, it is difficult to get biometric data such as heart rate and observe the body posture of the person. Therefore, we eliminate all these uncertainties by proposing a framework that needs only face visuals, audio and text data. Additionally, unlike other approaches, we leverage theories in psychology to develop our model design.

\section{Method}
\subsection{Proposed Model Design}
Since the patient population is individuals with mental illness, we used psychology and psychiatry literature to build our algorithm so that the recognition and understanding of engagement are as close as possible to a psychotherapist’s method of engagement evaluation during a session. We, therefore, take a multi-componential approach and propose a framework~\modelname (Multimodal Perception of Engagement for Telehealth) that estimates the engagement levels of the patient in terms of their affective and cognitive states. These modes (affective and cognitive) are basically the categories of the different cues used by mental health therapists to assess their patients. Additionally, since the extent to which patients remain engaged during the telehealth session is temporal in nature, we are interested in analyzing it across micro- level time scales in the range of a few seconds. These characteristics of our approach align perfectly with the person-oriented analysis discussed by~\cite{sinatra2015challenges}. 

\subsubsection{Cognitive State Mode}
The \textit{Cognitive state} involves comprehending complex concepts and issues and acquiring difficult skills. It conveys deep (rather than surface-level) processing of information whereby the person gains a critical or higher-order understanding of the subject matter and solves challenging problems. 

Psychotherapists usually measure and evaluate the cognitive state of the person using neuropsychological exams that are typically conducted via in-person interviews or self-evaluations to gauge memory, thinking, and the extent of understanding of the topic of discussion. There has been a lot of work around determining biomarkers for detecting signs of a person’s cognitive state. However, these methods are either offline or fail to consider various essential perceptual indicators. Recently, there has been a lot of work around using speech as a potential biomarker for detecting cognitive decline. For instance, stress negatively affects the cognitive functions of a person, and this can be easily detected using speech signals. Moreover, speech-based methods are attractive because they are non-intrusive, inexpensive, and can potentially be real-time. The following 4 audio features have proven to be extremely useful for checking signs of cognitive impairment and are increasingly being used to detect conditions such as Alzheimer's and Parkinson’s:– 
\begin{enumerate}
    \item \textit{Glottal features} ($f_g$) help in characterizing speech under stress. During periods of stress, there is an aberration in the amount of tension applied in the opening (abduction) and closing (adduction) of the vocal cords. 
	\item \textit{Prosody features} ($f_{pr}$) characterize the speaker’s intonation and speaking styles. Under this feature, we analyze variables like timing, intonation, and loudness during the production of speech. 
	\item \textit{Phonation} ($f_{ph}$) in people with cognitive decline is characterized by bowing and inadequate closure of vocal cords, which produce problems in stability and periodicity of the vibration. They are analyzed in terms of features related to perturbation measures such as jitter (temporal perturbations of the fundamental frequency), shimmer (temporal perturbation of the amplitude of the signal), amplitude perturbation quotient (APQ) and pitch perturbation quotient (PPQ). Apart from these, the degree of unvoiced is also included. 
	\item \textit{Articulation} ($f_{ar}$) is related to reduced amplitude and velocity of lip, tongue, and jaw movements. The analysis is based primarily on the first two vocal formants F1 and F2. 
\end{enumerate}
We, therefore, define features corresponding to cognitive state as the concatenation of these 4 audio features. Therefore, cognitive state features $h_c = concat (f_g, f_{pr}, f_{ph}, f_{ar})$

\subsubsection{Affective State Mode}
The \textit{Affective State} encompasses affective reactions such as excitement, boredom, curiosity, and anger. The range of affective expressions will vary based on individual demographic factors (e.g., age), cultural backgrounds/norms, and mental health symptoms. 
In order to understand the affective state, we check if there exists any inconsistency between the emotions perceived and the statement the person made.~\cite{balomenos2004emotion, porter2008reading} suggest that when different modalities are modeled and projected onto a common space, they should point to similar affective cues; otherwise, the incongruity suggests distraction, deception, etc. In other words, if $E1, E2$, and $E3$ represent the emotions perceived individually from what the patient said (text), the way they said it or sounded (audio) and how they looked/expressed (visuals) respectively, then the patient would be considered engaged if $E1, E2$, and $E3$ are similar otherwise they are disengaged.  Therefore, we adopt pretrained emotion recognition models to extract affective features corresponding to audio, visuals and text from each video sample separately:
\begin{enumerate}
    \item \textit{Audio} ($f_a$): Mel-frequency cepstrum (MFCC) features were extracted from the audio clips available in the data. The affective features were extracted using an MLP network that has been trained for emotion recognition in speech using the data available in the CREMA-D dataset. A feature vector was obtained corresponding to each audio clip. 
	\item \textit{Visuals} ($f_v$): The VGG-B architecture suggested in~\cite{arriaga2017real} was used to extract affective features from the video frames. The output dimensions of the second last layer were modified to give a feature vector of length 100. 
    \item \textit{Text} ($f_t$): We extract affect features from the text using a bert-based model that has been trained network on GoEmotions dataset. 
\end{enumerate}
We, therefore, represent the affective state of the patient as a concatenation of $f_a, f_v,$ and $f_t$. Hence, affective state features $h_a = concat(f_a, f_v, f_t)$

\subsubsection{Learning Network}
Obtaining a large amount of high-quality labeled data to train a robust model for predicting patient engagement is inevitably laborious and requires expert medical knowledge. Considering that unlabeled data is relatively easy to collect, we propose a semi-supervised learning-based solution. Semi-supervised learning (SSL) enables us to deploy machine learning systems in real-life applications (e.g., image search~\cite{lu2005semi}, speech analysis~\cite{yu2010active, liu2013graph}, natural language processing) where we have few labeled data samples and a lot of unlabeled data. There have also been some prior works that explore SSL to do engagement detection in non-medical domains. One of the earliest works in this direction includes~\cite{alyuz2016semi} where they consider the development of an engagement detection system, more specifically emotional or affective engagement of the student in a semi-supervised fashion to personalize systems like Intelligent Tutoring Systems according to their needs. ~\cite{nezami2017semi} conducted experiments to detect user engagement using a facial feature based semi-supervised model. Most state-of-the-art SSL methods use Generative Adversarial Nets (GANs)~\cite{goodfellow2014generative}. GANs are a class of machine learning models and typically have two neural networks competing with each other to generate more accurate predictions. These two neural networks are referred to as the generator and the discriminator. The generator's goal is to artificially manufacture outputs that could easily be mistaken as real data. The goal of the discriminator is to identify the real from the artificially generated data. In trying to generate high-quality outputs, the generator learns to capture the different possible variations in the input variables and therefore, the data manifold well. This is extremely helpful when we may not be able to access data containing a wide variety of similar engagement-related cues visible across different patients. We use a multimodal semi-supervised GAN-based network architecture to regress the values of an engagement corresponding to each feature tuple $h_T$. This improves our model’s generalizability and makes it more robust than the previously defined semi-supervised learning approaches. The network is similar to the semi-supervision framework SR-GAN proposed by~\cite{olmschenketal}. The main distinction as discussed is that unlike the original model, we train the generator to model the feature maps generated by the Cognitive and Affective state modules. The discriminator needs to distinguish between the true (labeled and unlabeled) feature maps with the corresponding fake feature maps and gives an estimate for engagement. However, we develop a generator to model the feature maps generated by Cognitive and Affective state modules ($h_T$). 4 loss functions are used to train this network – $L_{lab}, L_{un},L_{fake},L_{gen}$.
\begin{enumerate}
    \item \textit{Labeled Loss} ($L_{lab}$) : Mean squared error of model output ($\hat{y_t}$ ) with ground truth ($y_t$).
	\item \textit{Unlabeled Loss} ($L_{un}$): Minimize the distance between the unlabeled and labeled dataset’s feature space. 
	\item \textit{Fake Loss} ($L_{fake}$): Maximize the distance between unlabeled dataset’s features with respect to fake images. 
    \item \textit{Generator Loss} ($L_{gen}$): Minimize the distance between the feature space of fake and unlabeled data
\end{enumerate}
We also make use of a gradient penalty (P) to keep the gradient of the discriminator in check which helps convergence. The gradient penalty is calculated with respect to a randomly chosen point on the convex manifold connecting the unlabeled samples to the fake samples. 
The overall loss function used for training the network is 
\begin{equation}
    L = L_{lab}  + L_{un}  + L_{fake}   + L_{gen}  +\lambda P
\end{equation}

\begin{figure}[h]
    \centering
    \includegraphics[width=\textwidth]{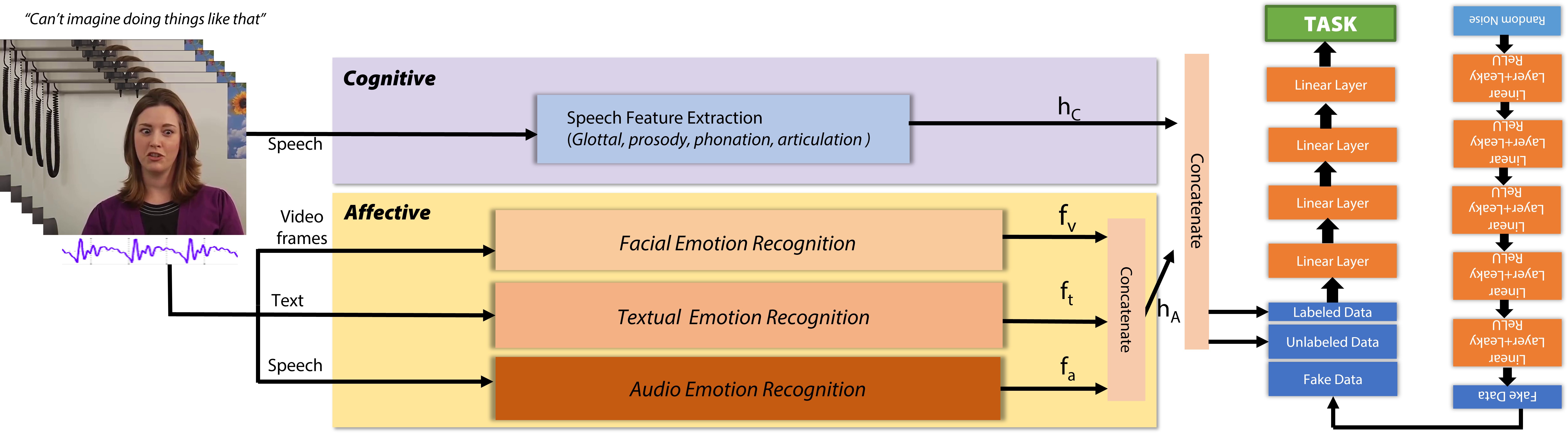}
    \caption{Overall block diagram of the proposed architecture. TASK here refers to the patient engagement estimation.}
\end{figure}

\section{Datasets}
\subsection{Multimodal Engagement Detection in Clinical Analysis (MEDICA)}
\begin{figure}[t]
    \centering
    \includegraphics[width=0.7\columnwidth]{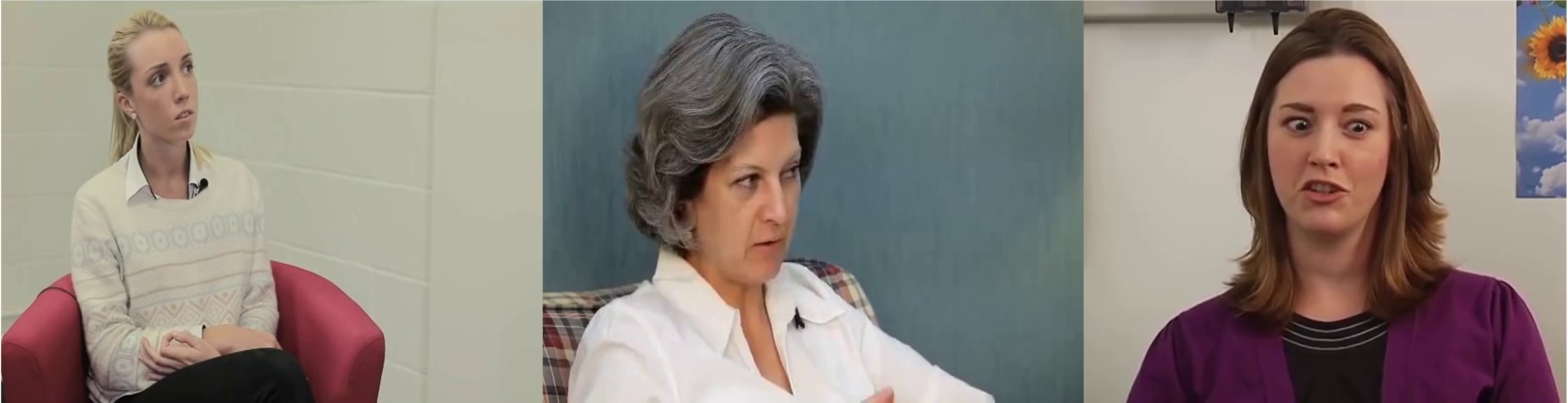}
    \caption{\small Examples from the MEDICA dataset created for mental health research. This dataset has been created using publicly available videos that are usually used for training purposes by different medical schools.}
    \label{fig:realworlddata}
\end{figure}
Engagement is an overloaded term, and the definition varies with the application, making it difficult and expensive to collect, annotate and analyze such data. As a result, we find too few multimodal-based engagement detection datasets currently available for us to use. Our problem statement revolves specifically around detecting patient engagement during a telemental health session. In such a setting, the only information we can work with includes the patient’s face and speech (audio and text). There exist datasets like CMU-MOSI~\cite{zadeh2016mosi}, CMU-MOSEI ~\cite{zadeh2018multimodal}, and SEND~\cite{ong2019modeling} that capture such settings. However, they are not specifically for engagement detection. Given the lack of a dataset that allows researchers to use multimodal features (video, text, and audio) for engagement, we propose MEDICA, a novel dataset developed specifically to cater to engagement detection using telemental health session videos. To use this data to address a broader range of issues related to mental health, we also include labels pertaining to stress and emotions. According to the author's knowledge, this dataset is one of the first publicly available datasets that caters specifically to multimodal research in patient engagement in mental health. Table \ref{tab:dataset} presents a comparison between MEDICA and other related datasets. 

Despite the rise in telehealth services and poor mental health patient-to-therapist ratios, there are no datasets that even try modeling telehealth sessions to give the community an opportunity to innovate and develop new technologies. 
MEDICA is a humble attempt by us to kick-start interesting research opportunities 
\paragraph{Acquisition:}
MEDICA has been developed by collecting publicly available mock therapy session videos created by different psychiatry medical schools for training their students. The patients in these videos are being advised for depression, social anxiety, and PTSD. We have collected 13 videos, each having a duration of around 20mins-30mins. We limit the videos to the setup wherein both the therapist and the patient are not visible together in the same frame. Additionally, we also take only those videos where there is only one patient. Each video has a unique English-speaking patient.

\paragraph{Processing and Annotation}
Since our only focus was to create a dataset that depicted the behavior of mental health patients during their sessions, we considered only parts of the videos where we had only the patient visible in the frames, which were scattered across the video for different durations. We took these scattered clips and divided them into smaller clips of 3 seconds each, resulting in a dataset of size 1229. We use Moviepy and speech-recognition libraries to extract audio and text from the video clips. Each video was annotated for attentiveness, stress, and engagement, which were scored on a Likert scale of [-3, 3]; hesitation was a binary target variable (Yes or No). Humans tend to have multiple emotions with varying intensities while expressing their thoughts and feelings. Therefore, the videos have been labeled for 8 emotions related to mental health: happy, sad, irritated, neutral, anxious, embarrassed, scared, and surprised. This will enable us to develop systems capable of understanding the various interacting emotions of the users. 

\begin{table}[h!]
\centering
\resizebox{\columnwidth}{!}{
\begin{tabular}{|c|c|c|c|c|c|p{2cm}|} 
\hline
   &  & &  &  &  & \\
\textbf{Dataset Name } & \textbf{Samples} & \textbf{Unique speakers} & \textbf{Modes}  & \textbf{Emotion}& \textbf{Engagement}& \textbf{Other mental health cues}\\
   &  & &  &  &  & \\
\hline
    &  & &  &  &  & \\
 RECOLA~\cite{ringeval2013introducing} &  3400 & 19  &\{v,a\}& \cmark& \xmark & physiological (electrocardiogram, and electrodermal activity)\\
 CMU MOSEAS~\cite{zadeh2020moseas} &715 & multiple& \{v,a,t\}& \cmark & \xmark & NA\\
 CMU MOSI~\cite{zadeh2016mosi} & 2199 & multiple & \{v,a,t\}& \cmark & \xmark & NA\\
 CMU MOSEI~\cite{zadeh2018multimodal} & 3228 & 1000 & \{v,a,t\} & \cmark & \xmark & NA\\
 SEND~\cite{ong2019modeling} & 193 & 49 & \{v,a,t\} & \cmark & \xmark & NA\\
 DAiSEE~\cite{gupta2016daisee}   &9068 & 112 & v & \xmark & \cmark &NA\\
 HBCU~\cite{whitehill2014faces} & 120  & 34 & v & \xmark & \cmark &NA\\
 in-the-wild~\cite{kaur2018prediction} &195 & 78  & v & \xmark & \cmark & NA  \\
 SDMATH~\cite{sathayanarayana2014towards} & 20 &20&\{v,a\} & \xmark & \cmark &NA\\
  &  & &  &  &  & \\
  \textbf{MEDICA}  & \textbf{1229*}    &\textbf{13} & \textbf{\{v,a,t\}} & \cmark & \cmark & \textbf{{hesitation, stress, attentiveness}}\\
\hline
\end{tabular}
}
\caption{\small{Comparison of the MEDICA dataset with other related datasets.  Modes indicate the subset of modalities present  from{(v)visual,(a)audio,(t)text}. *: Current status of the dataset. The size of the dataset will be increased.}}
\label{tab:dataset}
\end{table}

\subsection{Real-World Data}
\begin{figure}[t]
    \centering
    \includegraphics[width=0.7\columnwidth]{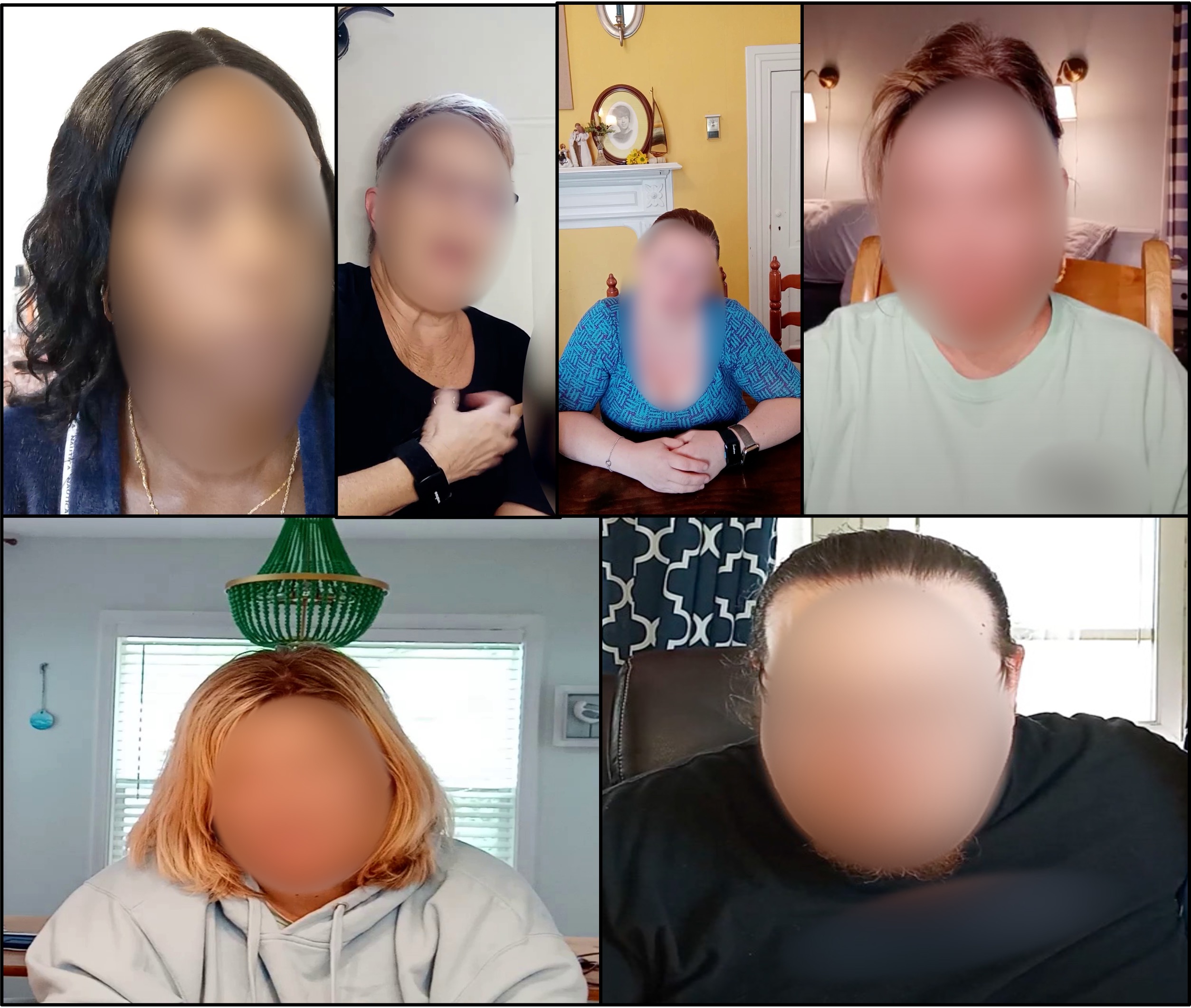}
    \caption{\small A few frames from the real-world videos we collected. The faces have been blurred here to protect the identity of the patients. However, the consent of the patients was taken to use their unblurred faces as input to ~\modelname.}
    \label{fig:realworlddata}
\end{figure}
We also wanted to test our methods in the real world and we collaborated with 8 child psychotherapists to do so. 20 caregivers voluntarily agreed to be part of this research after a psychotherapist explained to 
them its purpose and potential benefits, and that they could expect one or at max two of their telemental sessions to be recorded to test our proposed framework. They were also informed about the equipments that would be provided to them to ensure we get a clean recording. "Clean" refers to a recording executed with a camera of good quality with appropriate lighting conditions. The equipments mainly consisted of a smartphone having a good quality camera, a ring light with stand to ensure that the session was recorded in a well-lit environment, and internet connection to ensure that the session occurred smoothly without any network glitches. They were also given the assurance regarding preserving the confidentiality of the data being collected. The caregivers were informed that, during the video storage process, we would be “de-identifying” any facial images (using methods like blurring, etc) beside the caregivers, who may appear in the session video. We do this for two reasons. First, we are interested in only estimating the level of engagement of the caregivers and no other participant of the session. Second, the experiment is an agreement between only the caregiver and the therapist under the condition that no other person in the caregiver’s family (including the child) will be analyzed. Efforts were also made to limit their personal information, including limiting the experiment evaluations and medical records to only those people who are part of the study. On average, each of these sessions lasted around 20mins. The demographics of the caregivers who participated in our real-world experiments appear in Table\ref{caregiverdemograph}.  The entire data collection process can be divided into three parts:
\begin{enumerate}
    \item \textit{Pre-session}:  Before each telemental health session of a caregiver with their therapist, a research assistant helped the caregiver with set up the equipments to record their session. The assistant also ensured that the caregivers were comfortable using the equipment.
    \item \textit{During the session}: We ensured that the telemental health session ran just as it would normally. After the pre-session process, the research assistant would log off. Therefore, during the session, it would be just the therapist and the caregiver having a conversation. No one else from the study would be a part of it. The only thing different about this session was that the caregiver was being recorded using the smartphone given to them. We don’t record the therapist.
    \item \textit{Post-session}:  After the session was complete, a research assistant guided the participant regarding the steps to stop the recording and save the data collected. 
\end{enumerate}
\begin{table}[h!]
\centering
\resizebox{0.6\columnwidth}{!}{
\begin{tabular}{ |c|c|c|c|c| } 
\hline
\textbf{Measure} & \textbf{Item} & \textbf{Count} & \textbf{Percentage} \\
\hline
\multirow{3}{4em}{Gender} & Male & 1 & 5\% \\ 
& Female & 18 & 95\% \\
\hline
\multirow{5}{4em}{Race} & American Indian & 0 & 0\% \\
& Black American & 6 & 31\% \\
& Caucasian & 12 & 63\% \\
& Native Hawaiian & 1 & 5\% \\
& Biracial & 0 & 0\%\\
\hline 
\multirow{2}{4em}{Ethnicity} & Non-Hispanic & 19 & 100\%\\
& Hispanic & 0 & 0\% \\
\hline 
\multirow{6}{4em}{Household Income} & $ < \$20000$ & 2 & 10\%\\
& $ \$20000 - \$40000$ & 6 & 31\%\\
& $ \$40000 - \$60000 $ & 2 & 10\% \\
& $ \$60000 - \$80000 $ & 0 & 0\% \\
& $ \$80000 - \$100000 $ & 2 & 10\% \\
& $ > \$100000$ & 5 & 26\% \\
\hline
\end{tabular}
}
\caption{Demographic information for caregivers participating in the experiment}
\label{caregiverdemograph}
\end{table}
After a telemental health session is complete, the therapists score the collaborative relationship (therapeutic alliance) that was established between them and the caregiver during the session. The quality of this therapeutic alliance is measured using the working alliance inventory (WAI). WAI was modeled on Bordin’s theoretical work~\cite{bordin1979generalizability}. It captures 3 dimensions of the alliance – Bond, Task, and Goals. Extensive tests showed 12 items per dimension to be the minimum length for effective representations of the inventory. A composite score is computed based on these 12 items for each of the sessions conducted. Henceforth, we refer to this score as the WAI score. 

\section{Results and Discussion}
Motivated by recent works in clinical psychotherapy~\cite{bekes2021psychotherapists}, we use the standard evaluation metric of RMSE to evaluate our approach. 

\subsection{Study-1: Testing our proposed frameworks on MEDICA}
The purpose of the first study is to demonstrate the ability of our model to estimate the level of engagement exhibited by the patient in the video. This study was performed on the MEDICA dataset. 
As our proposed methodology leverages a semi-supervised approach, we extract labeled samples from MEDICA and unlabeled samples from the MOSEI dataset. After preprocessing, we extract 12854 unlabeled data points from MOSEI. We split the 1299 labeled data points from MEDICA into 70:10:20 for training, validation, and testing respectively. Therefore, the split of the labeled training data to unlabeled training data points is 909:12854. We compare our model with the following SOTA methods for engagement detection. 
\begin{enumerate}
    \item \textit{Kaur, Amanjot, et al} (LBP-TOP)~\cite{kaur2018prediction} use a deep multiple instance learning-based framework for detecting engagement in students. They extract LBP-TOP features from the facial video segments and perform linear regression using a DNN to estimate the engagement scores.
    \item \textit{Nezami, Omid Mohamad et al} (S3VM)~\cite{nezami2017semi} perform a semi-supervised engagement detection using a semi-supervised support vector machine.
\end{enumerate}
In addition to being SOTA, these methods can be used in a telehealth setting like ours. We use the publicly available implementation for LBP-TOP~\cite{kaur2018prediction}~and train the entire model on MEDICA. S3VM~\cite{nezami2017semi}~does not have a publicly available implementation. We reproduce the method to the best of our understanding.  

\begin{table}[h!]
\centering
\begin{tabular}{|l|c|}
\hline
\textbf{Method} & \textbf{RMSE for Engagement}\\\hline
LBP-TOP &  0.96\\ \hline
S3VM & 0.17  \\\hline
\textbf{MET (Ours)} & \textbf{0.10} \\
\hline
\end{tabular}
\vspace{5pt}
\caption{\small{Comparisons on~\dataname~ Dataset}}
\label{comparison_medica}
\end{table} 
Table~\ref{comparison_medica}~summarizes the RMSE values obtained for all the methods described above and ours. We observe an improvement of at least 40\%. Our approach is one of the first methods of engagement estimation built on the principles of psychotherapy. The modules used, specifically cognitive and affective states help the overall framework to effectively mimic the ways a psychotherapist perceives the patient’s level of engagement. Like psychotherapists, these modules also look for specific engagement-related cues exhibited by the patient in the video. 

\subsection{Study-2: Ablation Studies}
To show the importance of the different components (Affective and Cognitive) used in our approach, we run our method on MEDICA by removing either one of the modules corresponding to affective or cognitive state and report our findings. 
Table \ref{ablations_medica} summarizes the results obtained from the ablation experiments. We can observe that the ablated frameworks (i.e. only using affective (A) or cognitive (C) modules) do not perform as well as when we have both modules available. In order to understand and verify the contribution of these modules further, we leveraged the other labels (stress, hesitation, and attention) available in MEDICA and performed regression tasks using our proposed architecture on all of them. We observed that mode C performs better when predicting stress and hesitation values. Mode A performed better in estimating a patient’s level of attentiveness. These results agree with our understanding of cognitive state and affective state. Therefore, the combination of affective and cognitive state modes helps in efficiently predicting the engagement level of the patient.
\begin{table}[h!]
\centering
\resizebox{\columnwidth}{!}{%
\begin{tabular}{|c|c|c|c|c|c|c|} 
\hline
\textbf{Modality} & \textbf{RMSE for Engage} & \textbf{RMSE for Stress}& \textbf{RMSE for Hesitate}& \textbf{RMSE for Attention}\\\hline
A & 0.24 & 0.15 & 0.146 & 0.07 \\\hline
C & 0.3 & 0.13 & 0.16  & 0.08\\ \hline
A \& C & \textbf{0.10} & \textbf{0.12} & \textbf{0.14} & 0.1\\ \hline

\end{tabular}
}
\vspace{5pt}
\caption{\small{Ablation Experiments on MEDICA Dataset. We refer to Affective state mode by $A$ and Cognitive state mode by $C$.}}
\label{ablations_medica}
\end{table}
\subsection{Study-3:Analysis on Real-World Data}
MET trained for estimating engagement levels was tested on the processed real-world data. WAI scoring is based on certain observations the therapist makes during the session with the patient. The score obtained from our model is different than that from WAI, but we claim that like WAI, our estimates also capture the engagement levels of the patient well. If this is indeed the case, then both WAI and our estimates should be correlated. As discussed earlier, a single WAI score is reported by the therapist (provider) for the entire session. Since our framework performs microanalysis, we have engagement level estimates available for many instances during the session. Therefore, to make our comparison meaningful, we took the mean of the estimates obtained from MET for each session. We then observed the correlation between the mean scores of WAI and MET for the sessions. Instead of just taking the mean, we also took the median of the engagement level estimates available at different instances of the sessions and checked for their correlation with the WAI scores. Additionally, to quantify the quality of our framework's ability to capture the behavior of WAI, we performed the same correlation experiments with the comparison methods, S3VM and LBP-TOP frameworks. Table~\ref{comparison_corr_rwdata} shows the results of our experiments. Clearly, as compared to prior methods, our framework has been able to better understand WAI patterns and showcases a positive correlation.  

\begin{table}[h!]
\centering
\begin{tabular}{|l|*{2}{wc{\mylen}}|}
\hline
\multicolumn{1}{|c|}{Method} & \multicolumn{2}{c|}{Pearson Correlation Strength}\\ \hline
                             & \multicolumn{1}{c|}{Mean}          & {Median} \\ \hline
LBP-TOP                      & \multicolumn{1}{c|}{-0.03}         & {0}      \\ \hline
S3VM                         & \multicolumn{1}{c|}{-0.24}         & {-0.18}  \\ \hline
\textbf{MET (Ours)}          & \multicolumn{1}{c|}{\textbf{0.38}} & \textbf{0.40} \\ \hline
\end{tabular}
\caption{Correlation comparisons between different patient engagement estimates obtained from different methods and WAI for real-world data.}
\label{comparison_corr_rwdata}
\end{table}



The conceptual model of MET is also supported by Bordin’s 1979 theoretical work~\cite{bordin1979generalizability}. According to this theory, the therapist-provider alliance is driven by three factors – bond, agreement on goals, and agreement on tasks- and these factors fit nicely with the features identified in this work. While bond would correspond with affective, goals and task agreement correspond with cognitive. The merit of Bordin’s approach is that it has been used for child therapy and adults, and it is one of the more widely studied therapeutic alliance measures. Therefore, it is no surprise that our framework can work well to provide an estimate of engagement levels in a telemental health session.

\subsection{Conclusion}
Telehealth behavioral services that are delivered to homes via videoconferencing systems have become the most cost-effective, dependable, and secure option for mental health treatment, especially in recent times. Engagement is considered one of the key standards for mental health care. Given the difficulty in gauging the level of patient engagement during telehealth, an artificial intelligence-based approach has been shown to be promising for assisting psychotherapists.  We propose~\modelname, a novel multimodal semi-supervised GAN framework that leverages affective and cognitive features from the psychology literature to estimate useful psychological state indicators like engagement and valence-arousal of a person.  The method makes it possible to use the modalities easily available during a video call, namely, visuals, audio, and text to understand the audience, their reactions, and actions better. This can in turn help us have better social interactions. To the best of our knowledge, we are the first ones to do so. \modelname~can be an incredible asset for therapists during telemental health sessions. The lack of non-verbal cues and sensory data like heart rate makes it very difficult for them to make an accurate assessment of engagement (a critical mental health indicator). The lack of datasets has always been a big challenge to use AI to solve this and other mental-health-related tasks.  Therefore, to promote better research opportunities, we release a new dataset for engagement detection in mental health patients called~\dataname. We show our model’s usefulness on this as well as real-world data. As part of future work, we hope to build this dataset further to accommodate other related tasks apart from looking into possible kinds of variations arising due to cultural and geographical differences among patients and, therefore, making it more inclusive. 
Our work has some limitations and may not work well in case of occlusions, missing modality, and data corruptions due to low internet bandwidth. We plan to address this as part of future work. We would also like to explore making the predictions more explainable to enable psychotherapists to receive evidence-guided suggestions to make their final decisions. 

\bibliographystyle{ieeetr}
\bibliography{main}
\end{document}